\documentclass[12pt]{elsarticle}

\usepackage{threeparttable}
\usepackage{multicol}
\usepackage{geometry}
\usepackage{booktabs}
\usepackage{threeparttable}
\usepackage{graphicx}
\usepackage{multirow}%
\usepackage{amsmath,amssymb,amsfonts}%
\usepackage{amsthm}%
\usepackage{mathrsfs}%
\usepackage[title]{appendix}%
\usepackage{xcolor}%
\usepackage{textcomp}%
\usepackage{manyfoot}%
\usepackage{booktabs}%
\usepackage{algorithm}%
\usepackage{algorithmicx}%
\usepackage{algpseudocode}%
\usepackage{listings}%
\usepackage{float}
\usepackage{subfig}
\usepackage{type1cm}
\usepackage{siunitx}
\usepackage{algorithmicx}
\usepackage{lipsum}
\usepackage{wrapfig}
\usepackage{flushend}
\usepackage{adjustbox}
\usepackage{diagbox}
\usepackage{longtable}

\usepackage{hyperref}
\hypersetup{hidelinks,
	colorlinks=true,
	allcolors=black,
	pdfstartview=Fit,
	breaklinks=true}

\makeatletter
\newenvironment{breakablealgorithm}
{
		\begin{center}
			\refstepcounter{algorithm}
			\hrule height.8pt depth0pt \kern2pt
			\parskip 0pt
			\renewcommand{\caption}[2][\relax]{
				{\raggedright\textbf{\fname@algorithm~\thealgorithm} ##2\par}%
				\ifx\relax##1\relax 
				\addcontentsline{loa}{algorithm}{\protect\numberline{\thealgorithm}##2}%
				\else 
				\addcontentsline{loa}{algorithm}{\protect\numberline{\thealgorithm}##1}%
				\fi
				\kern2pt\hrule\kern2pt
			}
		}
		{
		\kern2pt\hrule\relax
	\end{center}
}
\makeatother

\journal{XXXX XXXX}
\begin{document}

\begin{frontmatter}



\title{ConsistencyTrack: A Robust Multi-Object Tracker with a Generation Strategy of Consistency Model}


\author[label1,label2]{Lifan Jiang}
\ead{lifanjiang@sdust.edu.cn,lifanjiang@zju.edu.cn}
\author[label1]{Zhihui Wang\corref{cor1}}
\ead{zhihuiwangjl@gmail.com}
\author[label1]{Siqi Yin}
\ead{siqiyin@sdust.edu.cn}
\author[label1]{Guangxiao Ma}
\ead{mgx@sdust.edu.cn}
\author[label1]{Peng Zhang}
\ead{pengzhang\_skd@sdust.edu.cn}
\author[label2]{Boxi Wu}
\ead{boxiwu@zju.edu.cn}
\cortext[cor1]{Corresponding author}
\affiliation[label1]{organization={College of Computer Science and Engineering, Shandong University of Science and Technology},
             city={Qingdao},
             country={China}}

\affiliation[label2]{organization={School of Software Technology, Zhejiang University},
            city={Ningbo},
            country={China}}

\begin{abstract}
Multi-object tracking (MOT) is a critical technology in computer vision, designed to detect multiple targets in video sequences and assign each target a unique ID per frame. Existed MOT methods excel at accurately tracking multiple objects in real-time across various scenarios. However, these methods still face challenges such as poor noise resistance and frequent ID switches. In this research, we propose a novel ConsistencyTrack, joint detection and tracking(JDT) framework that formulates detection and association as a denoising diffusion process on perturbed bounding boxes. This progressive denoising strategy significantly improves the model's noise resistance. During the training phase, paired object boxes within two adjacent frames are diffused from ground-truth boxes to a random distribution, and then the model learns to detect and track by reversing this process. In inference, the model refines randomly generated boxes into detection and tracking results through minimal denoising steps. ConsistencyTrack also introduces an innovative target association strategy to address target occlusion. Experiments on the MOT17 and DanceTrack datasets demonstrate that ConsistencyTrack outperforms other compared methods, especially better than DiffusionTrack in inference speed and other performance metrics. Our code is available at
\href{https://github.com/Tankowa/ConsistencyTrack}{https://github.com/Tankowa/ConsistencyTrack}.
\end{abstract}


\end{frontmatter}


\section{Introduction}\label{sec1}
Multi-object tracking(MOT) is a critical task in computer vision \cite{luo2021multiple,li2018deep,wang2024mm}, enabling real-time localization and tracking of specific targets' positions, sizes, and motion states within video sequences. Typical targets include various categories such as pedestrians, vehicles, or animals \cite{yang2023cooperative,liu2024prototype}. MOT algorithms take video sequences as input and output the targets' information, such as bounding boxes or trajectories.

Methodologies are primarily categorized into three paradigms: Tracking by detection (TBD) \cite{zhang2024scgtracker,qin2024towards,wojke2017simple}, joint learning of detection and embedding (JDE) \cite{wang2020towards,tsai2023swin}, and joint detection and tracking (JDT) \cite{kieritz2018joint,luo2024diffusiontrack}. The TBD paradigm begins with object detection followed by data association, with typical association strategies such as SORT \cite{bewley2016simple} and DeepSORT \cite{wojke2017simple}. SORT uses Kalman filtering and the Hungarian algorithm for tracking, while DeepSORT enhances performance with deep learning-based appearance features. However, a notable drawback of TBD is its reliance on the initial detection accuracy. If object detection fails, the subsequent tracking will likely be compromised, especially in complex scenes with occlusions or overlapping objects. Advancing technology introduced the JDE paradigm, integrating feature extraction into the detector to eliminate the need for separate re-identification modules. However, a limitation of JDE is that it sometimes compromises detection quality due to the joint optimization of detection and feature extraction, which may lead to reduced performance in both areas under complex scenarios. JDT paradigm emerges from advancements in technology, seamlessly combining the detection and tracking stages to enhance efficiency and minimize computational overlap. However, the integration of detection and tracking in JDT can lead to challenges in distinguishing between closely spaced objects and maintaining consistent track identities in dynamic environments, where objects interact frequently and the scene changes rapidly. Therefore, the tracking paradigms still should be innovated to improve their tracking abilities.

\begin{figure*}[!ht]
		\centering
		\includegraphics[width=0.8\textwidth]{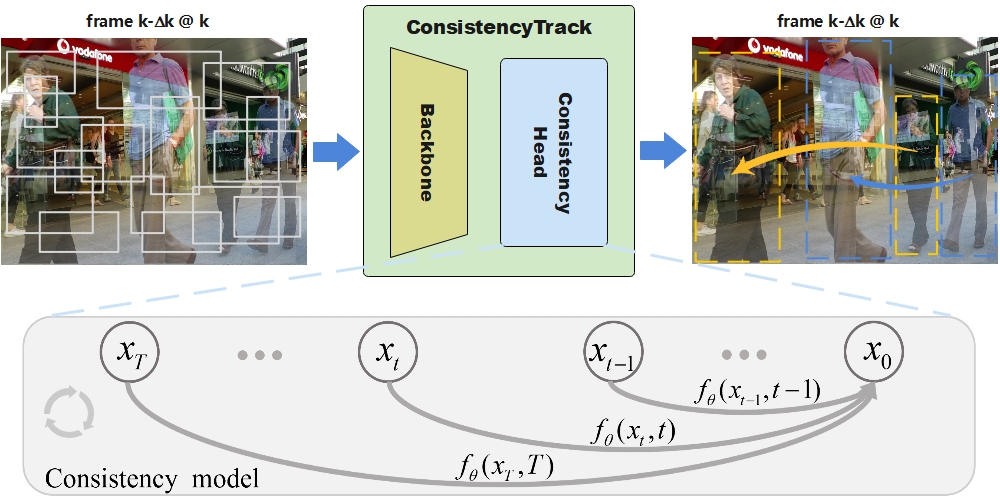}

		\caption{The denoising strategies of Consistency Model in the duty of MOT. ConsistencyTrack formulates object association as a denoising diffusion process from paired noise boxes to paired object boxes within two adjacent frames $(k-\Delta k, k)$. Here, $ f_{\theta}(\cdot,\cdot) $ represents a one-step denoising process.}
		\label{figmain1}
\end{figure*}

In recent years, diffusion models \cite{dhariwal2021diffusion,croitoru2023diffusion}, also known as score-based generative models, have demonstrated significant effectiveness in various domains, such as object detection \cite{chen2023diffusiondet}, image segmentation \cite{brempong2022denoising}, and image generation \cite{yuan2023efficient}. A defining characteristic of diffusion models is their iterative sampling mechanism, systematically reducing noise from initial random vectors, thereby greatly enhancing the model's robustness during the training phase. Building on the principles of diffusion models, DiffusionTrack \cite{luo2024diffusiontrack} has surpassed existing detectors in noise resistance, even exceeding the paradigm established by Transformer models \cite{meinhardt2022trackformer}. However, its gradual denoising process, opposite to its iterative noise addition, imposes limitations on flexibility and computational efficiency. To adapt this model for real-world applications, its iterative processes still need further optimized. To address this challenge, we propose an innovative approach with a generation strategy of Consistency Model, denoted as ConsistencyTrack. Different with the foundational concepts of DiffusionTrack, the denoising strategy employed by Consistency Model is illustrated in Fig. \ref{figmain1}. Notably, the self-consistency of Consistency Model allows denoising to be completed in a single step, significantly enhancing execution efficiency. Therefore, while maintaining detection accuracy, the number of denoising iterations can be significantly reduced.

\begin{figure*}[!ht]
		\centering
		\includegraphics[width=0.65\textwidth]{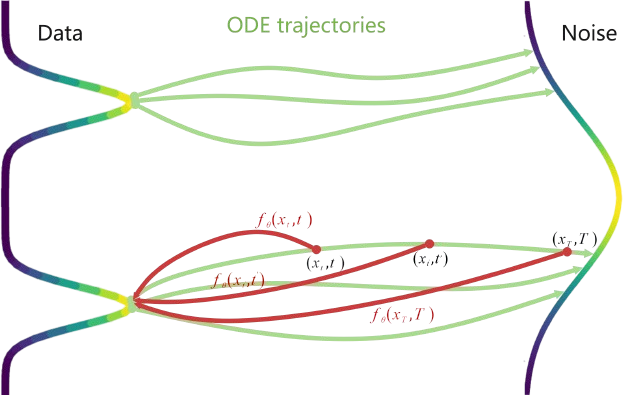}

		\caption{Consistency Model undergoes training process to establish a mapping that brings points along any trajectory of the PF ODE back to the origin of that trajectory \cite{song2023consistency}. The same as Fig. \ref{figmain1}, $ f_{\theta}(\cdot,\cdot) $ represents a one-step denoising process.}
		\label{ODE}
\end{figure*}

As illustrated in Fig. \ref{ODE}, our approach leverages the ordinary differential equation (ODE) framework for probability flow (PF), akin to the continuous-time model utilized in DiffusionTrack. These models effectively guide sample paths, facilitating a seamless transition from the initial data distribution to a manageable noise distribution. ConsistencyTrack distinguishes itself by mapping any given point from an arbitrary time step back to the origin of its trajectory. This is made possible by the model's self-consistency feature, ensuring that points along the same trajectory correspond to the same starting point. This innovative method enables the generation of data samples by transforming random noise vectors through a single network evaluation, starting from the starting points of ODE trajectories. The proposed ConsistencyTrack achieves efficient iterative sampling, distinct from the extremely low execution efficiency of DiffusionTrack, thereby improving the cost-effectiveness of the sampling process.

The proposed ConsistencyTrack innovatively integrates Gaussian noise into the center coordinates and dimensions extracted by the backbone network of bounding boxes, thereby generating corresponding noisy boxes. Subsequently, these generated noisy boxes are fed into a decoder for denoising prediction, primarily aligning them with ground truth (GT) boxes. It is noteworthy that to adapt ConsistencyTrack effectively to the JDT paradigm of MOT, two images at a fixed interval are simultaneously input into the network of ConsistencyTrack. This captures correlation information between instances of the same object across consecutive frames, thereby enhancing the model's ability for single-stage inference tracking.

Furthermore, we conducted rigorous evaluations of the proposed ConsistencyTrack's performance on the MOT17 and DanceTrack datasets. The experiments demonstrate that ConsistencyTrack exhibits robust noise resistance and fast inference speeds. Our work contributes to the field in the following aspects:

\begin{itemize}
\item ConsistencyTrack conceptualizes the process of object tracking as a generative denoising process and introduces a novel denoising paradigm. In contrast to the established paradigm in DiffusionTrack, which employs a very small number of iterations for noise addition and removal, our method represents a substantial advancement in enhancing the efficiency of the MOT task.

\item In crafting the loss function for the proposed ConsistencyTrack, we aggregate the individual loss values at time steps $(t-1, t)$ subsequent to the model’s predictions to compute the total loss. This methodology guarantees that the mapping of any pair of adjacent points along the temporal dimension to the axis origin maintains the highest degree of consistency. This attribute mirrors the inherent self-consistency principle central to Consistency Model.

\item We designed a novel target association strategy, distinct from DiffusionTrack within the JDT paradigm. This association strategy emphasizes the process of matching low-confidence detection boxes with tracking trajectories, significantly enhancing the ability to recognize occlusion issues and markedly improving performance metrics.
\end{itemize}

The structure of the paper is as follows: Section \ref{sec2} presents a concise review of the development of one-stage JDT methods and the application of traditional diffusion models in tracking tasks, and discusses the foundational principles of Consistency Model. Section \ref{sec3} delineates the specific methodologies for noise addition and removal within Consistency Model, elucidates the model’s architecture, and provides essential details regarding the training and sampling methodologies. Section \ref{sec4} details the empirical findings from evaluating ConsistencyTrack and conducts a comparative analysis against other leading models in the field. Finally, Section \ref{sec5} encapsulates the salient features of the newly proposed ConsistencyTrack and contemplates avenues for future research.

\section{Related works}\label{sec2}
\subsection{One-stage JDT methods}\label{subsec21}

In recent years, there have been several explorations into the one-stage paradigm, which combines object detection and data association into a single pipeline. Query-based methods, a burgeoning trend, utilize DETR \cite{zhu2020deformable} extensions for MOT by representing each object as a query regressed across various frames. Techniques such as TrackFormer \cite{meinhardt2022trackformer} perform simultaneous object detection and association using concatenated object and track queries. TransTrack \cite{sun2020transtrack} employs cyclical feature passing to aggregate embeddings, while MeMOT \cite{cai2022memot} encodes historical observations to preserve extensive spatiotemporal memory.

Offset-based methods, in contrast, bypass inter-frame association and instead focus on regressing past object locations to new positions. Examples include Tracktor++ \cite{cai2022memot} for temporal realignment of bounding boxes, CenterTrack \cite{zhou2020tracking} for object localization and offset prediction, and PermaTrack \cite{tokmakov2021learning}, which fuses historical memory to reason target location and occlusion. TransCenter \cite{xu2022transcenter} further advances this category by adopting dense representations with image-specific detection queries and tracking.

Trajectory-based methods extract spatial-temporal information from historical tracklets to associate objects. GTR \cite{zhou2022global} groups detections from consecutive frames into trajectories using trajectory queries, and TubeTK \cite{pang2020tubetk} extends bounding boxes to video-based bounding tubes for prediction. Both efficiently handle occlusion issues by utilizing long-term tracklet information.

\subsection{DiffusionTrack}\label{subsec22}
Diffusion models \cite{dhariwal2021diffusion,croitoru2023diffusion,chen2023diffusiondet,brempong2022denoising} originate from randomly distributed samples and progressively reconstruct the desired data through a denoising process. As powerful tools, these models have achieved significant success across a range of fields, including computer vision, natural language processing, and audio signal processing. In the task of MOT, diffusion models have been adopted into a tracking task known as DiffusionTrack \cite{luo2024diffusiontrack}. DiffusionTrack designs a novel tracker that performs tracking implicitly by predicting and associating the same object across two adjacent frames within the video sequence. This represents a groundbreaking application of Diffusion Model to the field of object detection. Building upon the foundations of DiffusionTrack, this work seeks to optimize the balance between detection accuracy and computational speed. We aim to enhance detection efficiency through a single-step processing approach, while preserving the essential benefits derived from the process of iterative sampling.

\subsection{Consistency Model}\label{subsec23}
Diffusion Model operates on an iterative generation process, which often results in limited execution efficiency, thereby restricting its applicability in real-time scenarios. To mitigate this limitation, OpenAI introduced Consistency Model, a novel class of generative models that can swiftly produce high-quality samples without the necessity for adversarial training. Consistency Model enables rapid one-step generation while also providing the flexibility for multi-step sampling to balance computational efficiency with the quality of generated samples. Additionally, it offers zero-shot data manipulation capabilities, including tasks such as image restoration, colorization, and super-resolution, without the need for task-specific training. This work formally acknowledges these capabilities and, for the first time, integrates Consistency Model in the field of MOT.

{\small
\begin{longtable}{p{2.2cm} p{10cm}}
\caption{Nomenclature with related notations.} \label{tab:test1_1} \\
\toprule
\textbf{Notation} & \textbf{Definition} \\
\endfirsthead
\multicolumn{2}{c}%
{{\tablename\ \thetable{} -- continued from previous page}} \\
\toprule
\textbf{Notation} & \textbf{Definition} \\
\endhead
\bottomrule
\endfoot
\bottomrule
\endlastfoot
$t_r$ & A random time step in the range $[0, T]$ \\
$t$ & Current time step \\
$T$ & Number of total time steps \\
$\Delta t$ & Time step interval for sampling \\
$k$ & The $k$-th frame in the video \\
$\Delta k$ & Time interval of the selected frames \\
$K$ & Number of total frames in the video \\
$u_{roi}^k$ & RoI-features at frame k \\
$q_{pro}^k$ & Self-attention output query \\
$q_k$ & The object query \\
$\textit{c}_x^\textit{i}/\textit{c}_y^\textit{i}$ & $x/y$-axis coordinate of the $i$-th box's center point \\
$\textit{w}^\textit{i}/ \textit{h}^\textit{i}$ & Width / Height of the $i$-th box \\
$B_i^k$ & $(\textit{c}_x^\textit{i}, \textit{c}_y^\textit{i}, \textit{w}^\textit{i}, \textit{h}^\textit{i})$ of the $i$-th box at frame k \\
$\alpha_t/\sigma_t$ & Parameter in Denoiser at the $t$-th time step \\
$\theta$ & Model parameter \\
$F_{\theta}(\cdot,\cdot)$ & A designed free-form deep neural network \\
$Split $ & Split function description \\
$f_{\mathrm{BMM}}(\cdot,\cdot)$ & Batch Matrix Multiplication function \\
$Linear(\cdot)$ & Fully-connected layer \\
$\mathcal{N}(\cdot,\cdot)$ & Normal distribution \\
$f_{\theta}(\cdot,\cdot)$ & Final answer for Consistency Model \\
$p_{\theta}(\cdot,\cdot) $ & Prediction function parameterized by $\theta$ \\
$c_{skip/out/in}(\cdot)$ & Calculation factor for $f_{\theta}$ \\
$\lambda(\cdot)$ & A positive weighting function \\
$\mathcal{L}$ & Total loss function in training phase \\
$\mathcal{L}_{cls/L1/GIoU3d}$ & Focal / L1 / GIoU3d loss item \\
$GIoU_{3d}(\cdot,\cdot)$ & Three-dimensional Generalized Intersection over Union \\
$\lambda_{cls/L1/GIoU3d}$ & Weight for Focal / L1 / GIoU3d loss item \\
$\sigma_{max/min}$ & Maximum / Minimum threshold of noise parameter \\
$\sigma_{data}$ & Noise parameter between $\sigma_{min}$ and $\sigma_{max}$ \\
$\epsilon$ & Randomly generated Gaussian noise \\
$B_t$ & Random noise at the $t$-th time step in sampling \\
$N $ & Batch size of concurrently processed samples \\
$R $ & Number of regions analyzed within each sample \\
$d $ & Dimension of feature \\
$r(\cdot)$ & Generate random noise with given dimensions \\
$E(\cdot)$ & Image feature extraction with backbone network \\
$P_c(\cdot,\cdot)$ & Prediction of Consistency Model in each time step \\
$nms(\cdot,\cdot)$ & Non-max suppression (NMS) operation \\
$N_{th}$ & Threshold of NMS operation \\
$B_{th}$ & Threshold of Box-renewal operation \\
$dcm(\cdot,\cdot)$ & Decoder of ConsistencyTrack with head network \\
$conc(\cdot,\cdot)$ & Concatenate function \\
$score(\cdot)$ & Estimation function of association score for each target \\
$frame(\cdot)$ & Number counter of adjacent untracked frames for each target \\
$n_{ss}$ & Number of sampling steps \\
$N_{train}$ & Number of total proposed boxes in training phase \\
$n_{rp}$ & Number of times the prior box repeats \\
$n_p$ & Number of total proposed boxes in inference \\
$n_r$ & Number of current proposed boxes \\
$x_s$ & Padded box information at time axis origin \\
$x_t$ & Noised box information at the $t$-th time step \\
$x_b$ & Predicted box information in each time step \\
$x_0$ & Predicted box information at time axis origin \\
$x_{box/cls/score}$ & Predicted box coordinate / category / association scores \\
AP & Average Precision \\

\end{longtable}
}

\subsection{Nomenclature}\label{subsec24}
For the sake of clarity in the ensuing discussion, we
provide a summary of the symbols and their corresponding descriptions as utilized in this study. This is encapsulated in Table \ref{tab:test1_1}, which meticulously outlines the nomenclature employed. The symbols encompass a variety of elements including training samples, components of the loss function, strategies for training, and metrics for evaluation, among others.

\section{The proposed tracking method}\label{sec3}

\begin{figure*}[!ht]
		\centering
		\includegraphics[width=1.0\textwidth]{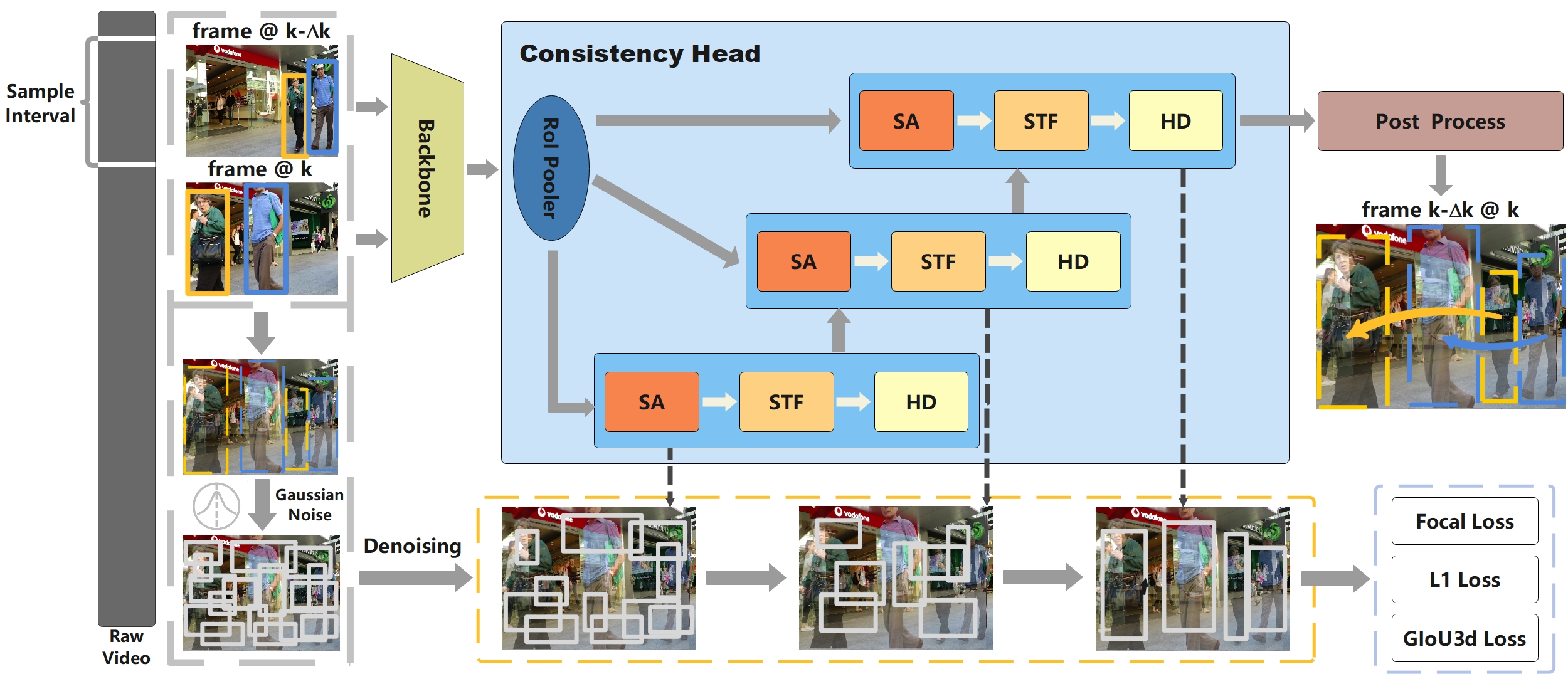}

		\caption{Training procedure of the proposed ConsistencyTrack. Features are extracted through the backbone network which extracts them from adjacent frames $(k-\Delta k,k)$ in a video sequence. Then, random Gaussian noise is added to the GT boxes according to the noise addition strategy of Consistency Model. These noisy boxes, with corresponding features, are processed by the RoI pooler and then input into the ConsistencyHead for iterative noise removal using three basic modules, ultimately yielding the final detection results. Each basic module contains a self-attention mechanism (SA), a Spatial-temporal fusion module (STF), and a correlation score head (HD). After the post process, the objects between adjacent frames $(k-\Delta k,k)$ are one-to-one associated with their matching scores.}
		\label{figmain}
\end{figure*}

In this section, we introduced the proposed tracking method with generation strategy of Consistency Model, denoted as ConsistencyTrack. This tracker is designed to perform the tracking duty implicitly by predicting and associating the same object across two adjacent frames within the video sequence. We first reviewed the pipeline of MOT, Diffusion Model, and Consistency Model. Finally, detailed discussions were provided on the training and inference procedures of the model.

\subsection{Preliminaries}

\textbf{Multi-object tracking.} The training samples of MOT are a set of input-target pairs \((X_k, B_k, C_k)\) per \(k\)-th frame, where \(X_k\) is the input image, \(B_k\) and \(C_k\) are a set of bounding boxes and ID information for objects in the video on the \(k\)-th frame respectively. More specifically, we formulate the \(i\)-th box in the set \(B_k\) as \(B_i^k = (c_i^x, c_i^y, w_i, h_i)\), where \((c_i^x, c_i^y)\) is the center coordinates of the bounding box and \( (w_i, h_i) \) are width and height of that bounding box, \(i\) is the identity number respectively. Specially, \(B_i^k = \emptyset\) when the \(i\)-th object is missing in \(X_k\).

\textbf{Diffusion Model.} Diffusion models \cite{dhariwal2021diffusion,croitoru2023diffusion,brempong2022denoising} emulate the image creation process through a sequence of stochastic diffusion steps. The core of diffusion models involves commencing with random noise and progressively refining it until it closely matches a sample from the target distribution. In the forward diffusion process, starting with a data point drawn from the real distribution, \( x_0 \sim q(x) \), Gaussian noise is incrementally introduced over \( T \) steps with the following iterative process:

\begin{equation}
q(x_t | x_{t-1}) = \mathcal{N} \left( x_t; \sqrt{1 - \beta_t} x_{t-1}, \beta_t I \right),
\end{equation}
where \( \beta_t \) schedules the noise for the current timestep \( t\in (1,T] \). In the reverse diffusion process, the random noise \( x_T \sim \mathcal{N}(0, I) \) is denoised into the target distribution by modeling \( q(x_{t-1}|x_t) \). At each reverse step \( t \), the conditional probability distribution is represented approximately by a network \( \epsilon_{\theta}(x_t, t) \) using the timestep \( t \) and the previous output \( x_t \) as input:

\begin{equation}
x_{t-1} \sim p_\theta(x_{t-1} | x_t) = \mathcal{N} \left( x_{t-1}; \frac{1}{\sqrt{\alpha_t}} \left( x_t - \frac{1 - \alpha_t}{\sqrt{1 - \bar{\alpha}_t}} \epsilon_\theta (x_t, t) \right), \beta_t I \right),
\end{equation}
where \( \alpha_t = 1 - \beta_t \) and \( \bar{\alpha}_t = \prod_{i=1}^t \alpha_i \). Through iterative operations, the noise in the current state is gradually reduced, eventually bringing it close to a real data point when approaching the original timestep of sample \( x_0 \).

\textbf{Consistency Model.}
Within the framework of Consistency Model which utilizes deep neural networks, two cost-effective methodologies are investigated for enforcing boundary conditions. Let \(F_{\theta}(x, t)\) represent a free-form deep neural network with the input \(x\). The first method directly parameterizes Consistency Model as:

\begin{equation}\label{func4}
f_{\theta}(x, t) =
\begin{cases}
  x, & \text{if } t = \tau, \\
  F_{\theta}(x, t), & \text{if } t \in (\tau, T),
\end{cases}
\end{equation}
where \(\tau\) is an integer in the range \([0, T-1]\). The second method parameterizes Consistency Model by incorporating skip connections and is formalized as follows:

\begin{equation}\label{func5}
f_{\theta}(x, t) = c_{skip}(t)x + c_{out}(t)F_{\theta}(x, t),
\end{equation}
where \(c_{skip}(t)\) and \(c_{out}(t)\) are differentiable functions \cite{song2023consistency}, satisfying \(c_{skip}(\tau) = 1\) and \(c_{out}(\tau) = 0\). By employing this construction, Consistency Model becomes differentiable at \(t = \tau\), provided that \(F_{\theta}(x, t)\), \(c_{skip}(t)\), and \(c_{out}(t)\) are all differentiable. This differentiability is crucial for the training process of continuous-time Consistency Models.

\subsection{Architecture}\label{subsec32}
The overall framework of our ConsistencyTrack is visualized in Fig. \ref{figmain}, which consists of two major components: a feature extraction backbone and a data association denoising head (diffusion head). The feature extraction backbone processes two adjacent input images \((X_{k-\Delta k}, X_k)\) to extract deep feature representations. The data association denoising head uses these features as conditions, rather than the raw images, to progressively refine paired association box predictions from paired noise boxes. In our setup, data samples consist of paired bounding boxes \(z_0 = (B_{k-\Delta k}, B_k)\), where \(z_0 \in \mathbb{R}^{N \times 8}\). A neural network \(f_\theta(z_s, s, X_{k-\Delta k}, X_k)\) for \(s = \{0, \cdots, K\}\) is trained to predict \(z_0\) from paired noise boxes \(z_s\), conditioned on the corresponding two adjacent images \((X_{k-\Delta k}, X_k)\). The corresponding association confidence score \(S\) are produced accordingly. If \(X_{k-\Delta k} = X_k\), the task of MOT degenerates into an object detection problem. This consistent property allows ConsistencyTrack to simultaneously solve both the tasks of object detection and tracking. It is noteworthy that $\Delta k$ is set to 1 only during the tracking of matching process, seen as Fig. \ref{figST}.

\textbf{Backbone.} We adopt the YOLOX backbone \cite{ge2021yolox}, which utilizes Feature Pyramid Networks (FPN) \cite{lin2017feature} to extract high-level features from two adjacent frames. These features are then fed into the diffusion head for the denoising process of conditioned data association.

\textbf{Diffusion head.} The diffusion head takes a set of proposal boxes as input to crop RoI features from the backbone's feature map. These RoI features are processed through different blocks to perform box regression, classification, and the prediction of association confidence score. To address the object tracking problem, each block of the diffusion head incorporates a Spatial-Temporal Fusion module (STF) and an association score head.

\textbf{Spatial-Temporal Fusion Module.} STF module is proposed to enable temporal information exchange between paired boxes across two consecutive frames, facilitating complete data association. Given the RoI features $\{u_{roi}^{k-\Delta k}, u_{roi}^k\} \in \mathbb{R}^{N \times R \times d}$ for two consecutive timesteps, where $N$ is the batch size, $R$ is the number of regions, $d$ is the feature dimension, and the self-attention output queries $\{q_{pro}^{k-\Delta k}, q_{pro}^k\} \in \mathbb{R}^{N \times d}$. Then, we proceed with the following transformations:

\begin{enumerate}
    \item \textbf{Linear Transformation and Splitting:}
    Each query $q_{pro}^i$ is first transformed using a linear projection, and the result is then split into two separate equidimensional tensors:
    \begin{equation}\label{func66}
    P_{1}^{i}, P_{2}^{i} = \textit{Split}(\textit{Linear}(q_{\text{pro}}^{i})).
    \end{equation}

    \item \textbf{Batch Matrix Multiplications:}
    RoI features from the two timesteps are concatenated and then subjected to two consecutive batch matrix multiplications (BMM) with the split parts:
    \begin{equation}\label{func77}
    \textit{feat} = f_{\mathrm{BMM}}\left(f_{\mathrm{BMM}}\left(\textit{conc}\left(u_{\text{roi}}^{i}, u_{\text{roi}}^{j}\right), P_{1}^{i}\right), P_{2}^{i}\right).
    \end{equation}

    \item \textbf{Final Linear Transformation:}
    The resulting feature tensor from the BMM operations is further processed using another linear transformation to produce the object queries for the current block:
    \begin{equation}\label{func88}
        q^{i} = \textit{Linear}(\textit{feat}), \quad q^{i} \in \mathbb{R}^{N \times d}.
    \end{equation}

    \item \textbf{Index Relationships:}
    The indices $(i, j)$ are taken from the adjacent time pairs $[(k-\Delta k, k), (k, k-\Delta k)]$, indicating the operation considers transitions between consecutive timesteps, in both forward and backward directions.
\end{enumerate}

\textbf{Association Score Head.} In addition to the box head and class head, we introduce an additional association score head. This head utilizes the fused features of paired boxes, obtained from the spatial-temporal fusion module, and feeds them into a linear layer. The output of this head provides the confidence score for data association. It determines whether the paired box outputs belong to the same object during the subsequent post-processing of NMS.

\subsection{Model Training}\label{subsec36}
During the training phase, the algorithm randomly selects a pair of frames from the video sequence as input to the model. First, GT boxes in the images are supplemented to a total number of $N_{\text{train}}$. Then, based on the noise addition strategy of Consistency Model, random noise is added to the original GT boxes in both frames. All these noised boxes are then fed into the model for the denoising process. Finally, model extracts the association relationships between the instance boxes in the two adjacent frames, calculates the loss, and performs the backpropagation operation. The detailed process of the training phase is described in Algorithm \ref{alg:3}.

\begin{breakablealgorithm}
\caption{Training loss of ConsistencyTrack}\label{alg:3}
\begin{algorithmic}[1]
\Require Images $(X_{k-\Delta k},X_{k})$ with GT boxes at two adjacent frames $(k-\Delta k,k)$
\Ensure Loss $\mathcal{L}_{t_r,t_{r-1}}$ per iteration
\For{each iteration}
    \State Sample $(X_{batch1},X_{batch2}) \in (X_{k-\Delta k},X_{k})$
    \State Extract features $E(X_{batch1},X_{batch2})$
    \State Pad $(X_{batch1},X_{batch2})$ with GT boxes and features as $x_s$
    \State Generate a random timestep $t_r\in [0, T]$

    \noindent{\color[RGB]{0,100,0}{\ \ \quad/* Calculate noise parameters */}}
    \State Calculate $(\sigma_{t_{r-1}}, \sigma_{t_r})$ by Eqn. (\ref{func6})
    \State Add noise to $x_s$ by Eqn. (\ref{func7}) as $x_{t_r}$

    \State Predict $x_{t_{r-1}}$ with $(x_{t_r}, \sigma_{t_r}, \sigma_{t_{r-1}}, x_s)$
    \State $d_{t_{r-1}} \gets dcm(x_{t_{r-1}}, \sigma_{t_{r-1}})$
    \State $d_{t_r} \gets dcm(x_{t_r}, \sigma_{t_r})$
    \State $\mathcal{L}_{t_r,t_r-1} \gets \mathcal{L}(d_{t_{r-1}},G) + \mathcal{L}(d_{t_r},G)$
    \State \Return Loss $\mathcal{L}_{t_r,t_{r-1}}$
\EndFor
\end{algorithmic}
\end{breakablealgorithm}
\hspace{0.5em}

\textbf{GT boxes padding.} In open-source benchmarks for MOT, as cited in \cite{hassan2024multi,islam2024position}, there is typically a variance in the number of annotated instances across images. To address this inconsistency, we implement a padding strategy by introducing auxiliary boxes around the GT boxes. This ensures that the total number of boxes reaches a predetermined amount, $N_{train}$, during the training phase. These padded instances are denoted as $x_s$, representing the original padded samples. For the $i$-th GT box, denoted as $b_i$, Gaussian noise is applied to its four parameters $(c^i_x, c^i_y, w^i, h^i)$ at a randomly selected timestep $t$.

\textbf{Box corruption.} The range of the noised box at the $t$-th timestep is constrained. Initially, the scale factor of the noise is determined as follows:

\begin{equation}\label{func6}
\sigma_{t} = \left(\sigma_{max}^{1/\rho} + \frac{t}{T - 1} \cdot \left(\sigma_{min}^{1/\rho} - \sigma_{max}^{1/\rho}\right)\right)^\rho.
\end{equation}

Subsequently, noise is introduced to the original padded sample $x_s$:

\begin{equation}\label{func7}
x_t = x_s + \epsilon \cdot \sigma_{t},
\end{equation}
where $\epsilon$ denotes randomly generated Gaussian noise.
Finally, the range of the noised box is restricted by:

\begin{equation}\label{fun9}
x_t \gets \frac{c_{in}(t)}{2} \cdot x_t,
\end{equation}
where $c_{in}(\cdot)$ represents the scale factor of the noised box and is defined as:

\begin{equation}\label{fun8}
c_{in}(t) = \frac{1}{\sqrt{{\sigma_{t}}^2 + \sigma_{data}^2}}.
\end{equation}

This formulation ensures that the noise scale factor is properly adjusted across the time steps, and that the noised box remains within the specified value ranges.

\begin{figure*}[!ht]
		\centering
		\includegraphics[width=0.75 \textwidth]{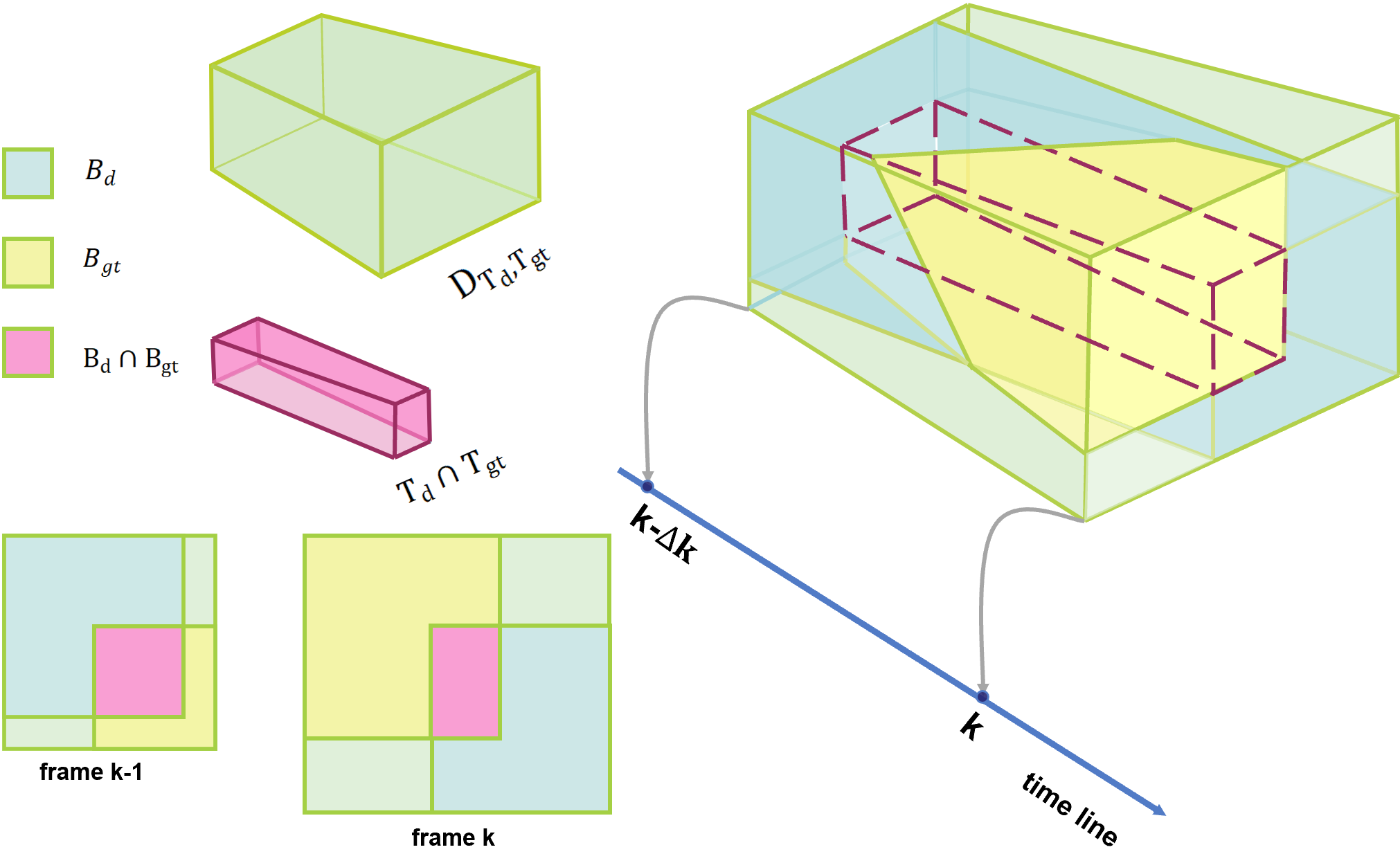}
		\caption{Visualization of the computation methodology for 3D GIoU. The volumetric intersection and the minimal bounding volume between target representations across consecutive frames are characterized as square frustums.
}
		\label{giou}
\end{figure*}

\textbf{Loss Function.} The loss function used to evaluate the predicted bounding boxes follows the framework established by DiffusionTrack, which incorporates both $\mathcal{L}_{\text{L1}}$ loss and $\mathcal{L}_{\text{GIoU3d}}$ loss item. The former represents the standard L1 loss item, while the latter represents the Generalized Intersection over Union (GIoU) loss. Notably, we extend the definition of GIoU to make it compatible with the paired boxes design. 3D GIoU and 3D IoU are the volume-extended versions of the original area-based ones. Additionally, focal loss $\mathcal{L}_{cls}$ is used to evaluate the classification of each predicted bounding box. To balance the relative impact of each loss component, a positive real-valued weight $\lambda_{\text{cls}/\text{L1}/\text{GIoU3d}}\in \mathbb{R}^+$ is assigned to each loss item. Therefore, the total loss function is formulated at the $t$-th timestep as follows:

\begin{equation}\label{fun10}
\mathcal{L}_t = \lambda_{cls} \cdot \mathcal{L}_{{cls}_t} + \lambda_{L1} \cdot \mathcal{L}_{{L1}_t} + \lambda_{GIoU3d} \cdot \mathcal{L}_{{GIoU3d}_t},
\end{equation}

with

\begin{equation}\label{fun11}
\mathcal{L}_{GIoU3d} = 1 - GIoU_{3d}(T_d,T_{gt}),
\end{equation}
where $T_d$ and $T_{gt}$ are square frustums consisting of estimated detection boxes and ground-truth bounding boxes for the same target in two adjacent frames respectively.

As shown in Fig. \ref{giou}, 3D GIoU of paired predicted boxes is defined as:

\begin{equation}\label{w}
\begin{aligned}
GIOU_{3D}(T_d, T_{gt}) = & - \frac{\left| \sum_{i=k-1}^{k} \left( Area(D_{B_d^i,B_{gt}^i}) - Area(B_d^i \cup B_{gt}^i) \right) \right|}{\left| \sum_{i=k-1}^{k} Area(D_{B_d^i,B_{gt}^i}) \right|} \\
& + IOU_{3D}(T_d, T_{gt}),
\end{aligned}
\end{equation}
where \(D_{B_d^i,B_{gt}^i}\) represents the smallest convex hull that includes the estimated detection box \(B_d\) and the ground-truth bounding box \(B_{gt}\) at the $i$-th frame. \(T_d\) and \(T_{gt}\) are similar to Eqn.(\ref{fun11}). The intersection \(T_d \cap T_{gt}\) also forms a square frustum, encompassing the overlaps \(B_{d}^{k-1} \cap B_{gt}^{k-1}\) and \(B_{d}^k \cap B_{gt}^k\).

Leveraging the self-consistency property of Consistency Model, the perturbed bounding boxes associated with sample $x_s$ at consecutive timesteps $t-1$ and $t$ undergo a joint denoising process. The corresponding loss values are accumulated to derive the final comprehensive loss:

\begin{equation}
\begin{aligned}
\mathcal{L} =& \lambda_{cls} \cdot (\mathcal{L}_{{cls}_{t-1}} + \mathcal{L}_{{cls}_t}) + \lambda_{L1} \cdot (\mathcal{L}_{L1_{t-1}} + \mathcal{L}_{L1_t}) \\
&+ \lambda_{GIoU3d} \cdot (\mathcal{L}_{{GIoU3d}_{t-1}} + \mathcal{L}_{{GIoU3d}_t}).
\end{aligned}
\end{equation}

\subsection{Inference}\label{subsec37}

\begin{figure*}[!ht]
		\centering
		\includegraphics[width=1.0 \textwidth]{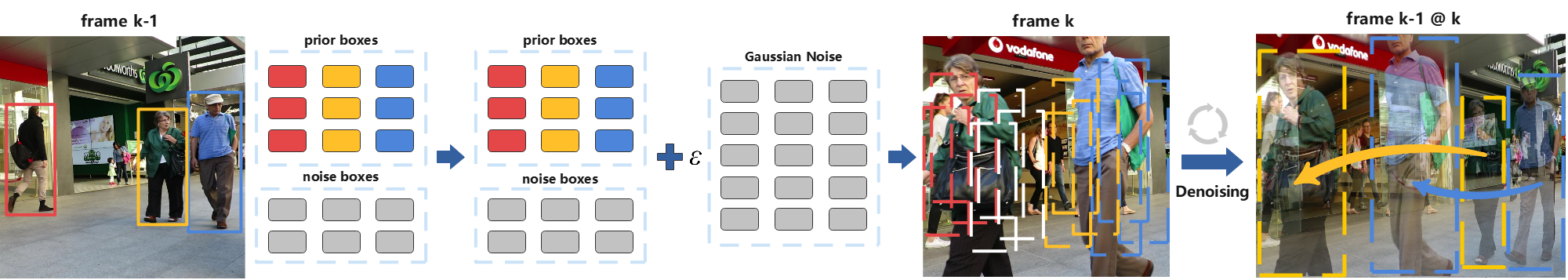}
		\caption{Illustration of the inference process with Consistency Model. First, padding repeated prior boxes with Gaussian boxes until the predefined number $N_{\text{test}}$ is reached. Then, adding Gaussian noise to the input boxes according to $x_t = x_s + \epsilon \cdot \sigma_{t}$ under the control of $\epsilon$. Finally, obtaining tracking results through a denoising process with one sampling step of Consistency Model.
}
		\label{figST}
\end{figure*}

The inference mechanism employed in ConsistencyTrack resembles that of DiffusionTrack, utilizing a denoising sampling strategy that starts from initial bounding boxes, similar to the processed noisy samples during the training phase, and progresses to the final object detections. In the absence of GT annotations, these initial bounding boxes are randomly generated following a Gaussian distribution. The model iteratively refines these predictions through multiple sampling steps. Ultimately, the final detections include refined bounding boxes and category classifications. After completing all iterative sampling steps, the predictions undergo enhancement through a post-processing module, resulting in the final outcomes. The detailed procedure is outlined in Algorithm \ref{alg:5}. Regarding the detailed demonstration of the inference phase, refer to Fig. \ref{figST}. Intuitive comparison of ConsistencyTrack and DiffusionTrack during the inference phase, as illustrated in the following Fig. \ref{com1}.

\begin{breakablealgorithm}
    \caption{Inference of ConsistencyTrack}
    \label{alg:5}
    \begin{algorithmic}[1]
        \Require Images $(X_{k-\Delta k}, X_{k})$ at the frame pair $(k-\Delta k, k)$, total timestep $T$, the number of sampling steps $n_{ss}$
        \Ensure Final predictions $nms(x_{box}, x_{cls}, x_{score})$

        \noindent{\color[RGB]{0,100,0}{/* Initialization */}}
        \State $\Delta t = T / n_{ss}$
        \State Generate random noise $B_0$ with the dimensions of presupposed boxes' amount
        \State Extract features $E(X_{k-\Delta k}, X_{k})$

        \noindent{\color[RGB]{0,100,0}{/* Iterative operation */}}
        \For{$t = 0$ \textbf{to} $T-1$ \textbf{step} $\Delta t$}
            \State Calculate ${\sigma}_t$ by Eqn. (\ref{func6})
            \State $x_0, x_b, x_{cls}, x_{box}, x_{score} \gets P_c(E(X_{k-\Delta k}, X_{k}), B_t)$
            \State Perform Box-renewal operation for $x_b$ and $x_0$
            \State $\nabla_\sigma x \gets (x_b - x_0) / {\sigma}_t$
            \State $B_t \gets x_b + \nabla_\sigma x (\sigma_{t + \Delta t} - \sigma_t)$

            \noindent{\color[RGB]{0,100,0}{\;\ \quad/*Supplement new proposals */}}
            \State $B_t \gets \text{conc}(B_t, r([1, n_p - n_r, 4]) \cdot \sigma_{t + \Delta t})$
        \EndFor


        \State \Return $nms(x_{box}, x_{cls}, x_{score})$
    \end{algorithmic}
\end{breakablealgorithm}

\subsection{Target association strategy}\label{subsec38}

Our target association process adopts the JDT paradigm, and the entire object matching process no longer involves additional feature matching. Initially, paired detection boxes are filtered based on the association scores obtained by the detector, retaining only those detection boxes with higher association scores. This is a specific embodiment of the JDT paradigm. Subsequently, the detection boxes are classified into high and low confidence for separate tracking matching. To address potential occlusions, a simple Kalman filter is implemented to reassociate lost objects. The pseudo-code of ConsistencyTrack is listed in Algorithm \ref{alg:ConsistencyTrack}.

\begin{breakablealgorithm}
\caption{ConsistencyTrack}
\label{alg:ConsistencyTrack}
\begin{algorithmic}[1]
\Require Video sequence $V$, a single track $m$, consistency track $CT$, association score threshold $\tau_{conf}$, detection score threshold $\tau_{det}$, track score threshold $\tau_{track}$, number of boxes for association $N_a$, the upper threshold $n_{lost}$ of adjacent frame amounts before lost, the high / low confidence detection boxes detected from the first half of the input information $D_{lpre / spre}$ and the second half of the input information $D_{lcur / scur}$, the intermediate parameter for returned collection $T_{act\_remain / lost\_remain / rm\_remain}$, trajectories with lost tags in this tracking match $T_{lost\_remain}$, trajectories with remove tags in this tracking match $T_{rm\_remain}$, trajectories waiting to be activated and already activated trajectories being updated during this tracking match $T_{act\_remain}$.
\Ensure Tracked targets' status $T_{activated / unactivated / lost / remove}$ per frame

\For{frame $(u_{k-1}, u_k)$ in $V$}
    \State $D_k \gets CT(u_{k-1}, u_k)$
    \State $D_{pre}, D_{cur}, D_{new} \gets \emptyset$
    \For{$(idx, d_{k-1}, d_k)$ in $D_k$}
        \If{$score(d_k) > \tau_{conf}$}
            \If{$idx < N_a$}
                \State $D_{pre} \gets D_{pre} \cup \{d_{k-1}\}$
                \State $D_{cur} \gets D_{cur} \cup \{d_k\}$
            \Else
                \State $D_{new} \gets D_{new} \cup \{d_{k-1}, d_k\}$
            \EndIf
        \EndIf
    \EndFor

    \noindent{\color[RGB]{0,100,0}{\;\ \quad/* Partitioning detection boxes */}}
    \For{$(idx, d_{k-1}, d_{k})$ in $D_{pre}, D_{cur}$}
        \If{$score(d_{k-1}) < \tau_{track} \ \&\ score(d_{k-1}) > 0.1$}
            \State $D_{spre} \gets d_{k-1}$
        \EndIf
        \If{$score(d_{k-1}) > \tau_{track}$}
            \State $D_{lpre} \gets d_{k-1}$
        \EndIf
        \If{$score(d_{k}) < \tau_{track} \ \&\ score(d_{k}) > 0.1$}
            \State $D_{scur} \gets d_{k}$
        \EndIf
        \If{$score(d_{k-1}) > \tau_{track}$}
            \State $D_{lcur} \gets d_{k}$
        \EndIf
    \EndFor

    \For{$(idx, d_{k-1}, d_{k})$ in $D_{new}$}
        \If{$score(d_{k-1}) \in (0.1,\tau_{track}) \ \&\  score(d_{k}) \in (0.1, \tau_{track})$}
            \State $D_{snew} \gets \{d_{k-1}, d_k\}$
        \EndIf
        \If{$score(d_{k-1}) > \tau_{track} \ \&\ score(d_{k}) > \tau_{track}$}
            \State $D_{lnew} \gets \{d_{k-1}, d_k\}$
        \EndIf
    \EndFor

    \noindent{\color[RGB]{0,100,0}{\;\ \quad/* Partitioning tracking trajectory */}}
    \State$T_{lost} \gets \{m \in T_{activated} \mid m \text{ is lost}\}$
    \State $T_{non\_lost} \gets \{m \in T_{activated} \mid m \text{ is not lost}\}$

    \noindent{\color[RGB]{0,100,0}{\;\ \quad/* First target association*/}}
    \State Associate $T_{non\_lost}$ and $D_{lpre}$ using IoU Similarity
    \State $U_{track} \gets \{m \in T_{non\_lost} \mid m \text{ not matched}\}$
    \State $U_{detection} \gets \{d \in D_{pre} \mid d \text{ not matched}\}$
    \State $T_{act} \gets \{m \in T_{non\_lost} \mid m \text{ matched}\}$
    \State \text{Associate} $T_{act}$ and $D_{lnew}$ \text{ using IoU Similarity}
    \State $U_{lnew} \gets \{d \in D_{lnew} \mid d \text{ not matched with } T_{act}\}$

    \State $T_{act\_remain} \gets \{m \in T_{act}\}$

    \noindent{\color[RGB]{0,100,0}{\;\ \quad/* Second target association*/}}
    \State Associate $T_{lost}$ and $U_{lnew}$ using IoU Similarity
    \State $T_{act\_remain} \gets \{m \in T \mid m \text{ matched with } U_{lnew}\}$
    \State $T_{unactivated} \gets \{d \in U_{lnew} \mid d \text{ unmatched with } T_{lost}\}$
    \State $U_{track\_now} \gets \{m \in T \mid m \text{ unmatched with } U_{lnew}\}$
    \State $T_{rm\_remain} \gets \{m \in T \mid m \text{ unmatched with } U_{lnew} \text{ and } frame(m) > n_{lost}\}$

    \noindent{\color[RGB]{0,100,0}{\;\ \quad/* Third target association*/}}

    \State Associate $U_{track}$ and $D_{sper}$ using IoU Similarity
    \State $U_{track} \gets \{m \in U_{track} \mid m \text{ not matched}\}$
    \State $U_{detection} \gets \{d \in D_{sper} \mid d \text{ not matched}\}$
    \State $T_{act} \gets \{m \in U_{track} \mid m \text{ matched}\}$
    \State \text{Associate} $T_{act}$ and $D_{snew}$ \text{ using IoU Similarity}
    \State $D_{scur} \gets \{D_{snew} \mid d \text{ matched with } T_{act}\}$
    \State $T_{act\_remain} \gets \{m \in T_{act}\}$

    \State $T_{lost\_remain} \gets \{m \in T \mid m \text{ not matched with } U_{track}\}$

    \noindent{\color[RGB]{0,100,0}{\;\ \quad/* Fourth target association*/}}
    \State Associate $T_{unactivated}$ and $U_{detection}$ using IoU Similarity
    \State $T_{act\_remain} \gets \{m \in T \mid m \text{ matched with } U_{detection} \text{ and } m \text{ is activated}\} $
    \State $T_{refind} \gets \{m \in T \mid m \text{ matched with } U_{detection} \text{ and } m \text{ is not activated}\} $
    \State $T_{unactivated} \gets \{d \in U_{detection} \mid d \text{ unmatched with } T_{unactivated}\}$

    \noindent{\color[RGB]{0,100,0}{\;\ \quad/* Update tracking status*/}}
    \State $T_{rm\_remain} \gets \{m \in T_{lost} \mid frame(m) > n_{lost}\}$

    \State $T_{activated} \gets T_{activated} \cup T_{refind} \cup T_{act\_remain}$
    \State $T_{lost} \gets ((T_{lost} \setminus T_{refind}) \cup T_{lost\_remain}) \setminus T_{rm\_remain} $
    \State $T_{remove} \gets T_{rm\_remain}$

\State \Return $T_{activated},T_{unactivated},T_{lost},T_{remove}$
\EndFor
\end{algorithmic}
\end{breakablealgorithm}

\section{Experiments}\label{sec4}
In this section, the performance of the proposed ConsistencyTrack is evaluated on two popular datasets: MOT17 and DanceTrack \cite{shan2020tracklets,zhang2023motrv2,insafutdinov2017arttrack,sun2022dancetrack}. Firstly, the noise robustness and various characteristics of ConsistencyTrack is tested through experiments. Then, the proposed ConsistencyTrack framework is compared with a series of established MOT models on several evaluation indicators. Finally, ablation studies are conducted to compare the optimal parameters of the ConsistencyTrack model, highlighting the significant efficiency advantages of our proposed model over DiffusionTrack.

\textbf{MOT17 Dataset.} The MOT17 dataset \cite{shan2020tracklets,zhang2023motrv2} is a widely used benchmark for MOT tasks, consisting of 14 challenging video sequences from various indoor and outdoor environments. It provides detailed annotations for object bounding boxes and identities, allowing for the evaluation of tracking accuracy using metrics like MOTA and MOTP. The dataset is known for its diversity and complexity, including scenarios with occlusions, varying lighting conditions, and dense crowds.

\textbf{DanceTrack Dataset.} The DanceTrack dataset \cite{insafutdinov2017arttrack,sun2022dancetrack} is designed to evaluate tracking algorithms in dynamic and complex scenarios, specifically focusing on dance performances. It includes sequences with fast, non-linear movements and frequent occlusions. The dataset provides detailed annotations for dancers, including bounding boxes and identities, and uses standard tracking metrics like MOTA and MOTP to assess performance. DanceTrack is particularly challenging due to the high-speed and intricate interactions among dancers.

\subsection{Implementation details}\label{subsec41}
We adopted the pre-trained YOLOX detector from ByteTrack \cite{luo2024diffusiontrack}, and trained ConsistencyTrack on the training sets of MOT17 and DanceTrack separately. For MOT17, the training schedule contains 60 training epochs of detection on the combined datasets (includes MOT17, CrowdHuman, Cityperson, and ETHZ), and 60 training epochs solely on MOT17 for tracking. For DanceTrack, no additional training data were used, and the model was trained by 80 epochs. During the detection and tracking training phases, we also employed data augmentation techniques such as Mosaic \cite{bochkovskiy2020yolov4} and Mixup \cite{zhang2017mixup}. Each training sample (frame pair) was directly sampled within each video with a frame interval of $\Delta k=5$. The input image size was resized to 1440×800. The AdamW optimizer \cite{loshchilov2017decoupled} was used with an initial learning rate of 1e-4, which decreased according to a cosine function with a final reduction factor of 0.1. We used a warm-up learning rate of 2.5e-5 with a warm-up factor of 0.2 for the first epoch. The model was trained on a single NVIDIA GeForce RTX A100 GPU with FP32 precision and a constant seed for all experiments. The mini-batch size was set to 3, with each GPU hosting two batches with $N_{train} = 500$. Our approach was implemented in Python 3.8 with PyTorch 1.10.

\subsection{Main Properties}\label{subsec42}
The core characteristic of ConsistencyTrack is its self-consistency, which ensures that the mapping effect from any point along the time axis back to the origin remains relatively stable. This stability indicates that once the model is sufficiently trained, it can quickly obtain inference results in very few sampling time steps. In addition, the addition of noise during the training process significantly improved the model's robustness to noise. Therefore, in different situations, the model can adjust the number of noise boxes according to specific requirements, thereby balancing accuracy and algorithm efficiency.

\begin{figure*}[!ht]
		\centering
		\includegraphics[width=0.7 \textwidth]{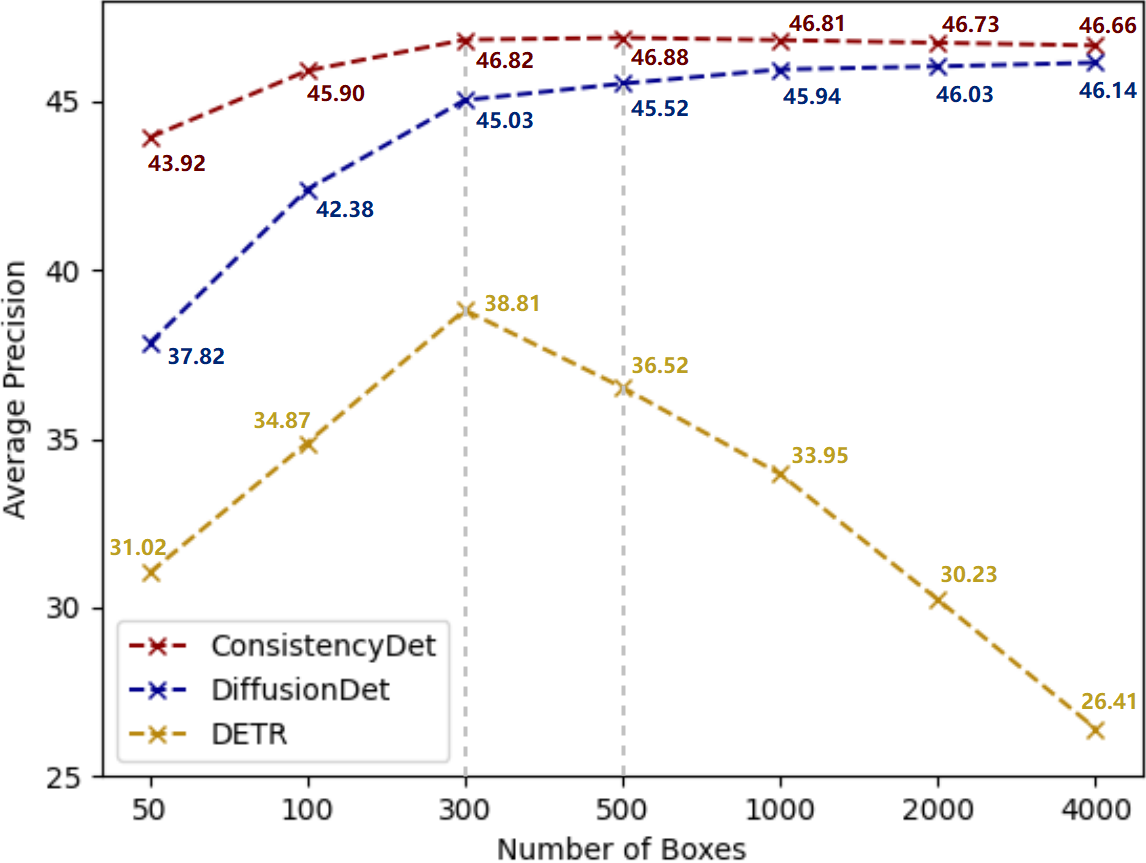}
		\caption{Performance comparisons of ConsistencyDet and DETR on COCO val dataset with increasing number of noisy boxes \cite{jiang2024consistencydet}.}
		\label{dmy}
\end{figure*}

\textbf{Robustness to detection perturbation.} To rigorously assess the robustness of ConsistencyTrack to noise during the detection phase, we independently trained and tested its detection component, ConsistencyDet \cite{jiang2024consistencydet}, on the MS-COCO dataset. This evaluation was benchmarked against leading detectors such as DiffusionDet and DETR, whose performance metrics are derived from \cite{chen2023diffusiondet}. As depicted in Fig. \ref{dmy}, ConsistencyDet exhibits a marked enhancement in performance correlating with the incremental inclusion of bounding boxes, achieving stability at $n_p > 300$ and peaking at $n_p = 500$. In contrast, DETR reaches its maximum AP at $n_p = 300$, thereafter experiencing a precipitous decline, notably decreasing to 26.4\% at $n_p = 4000$, a 12.4\% drop from its highest AP of 38.8\%. Although DiffusionDet also improves as more boxes are considered, its performance consistently trails behind that of ConsistencyDet. Consequently, ConsistencyDet not only proves more robust against noise but also showcases superior transferability and generalizability across diverse scenarios involving variable object counts.

\begin{figure*}[!ht]
		\centering
		\includegraphics[width=1.0 \textwidth]{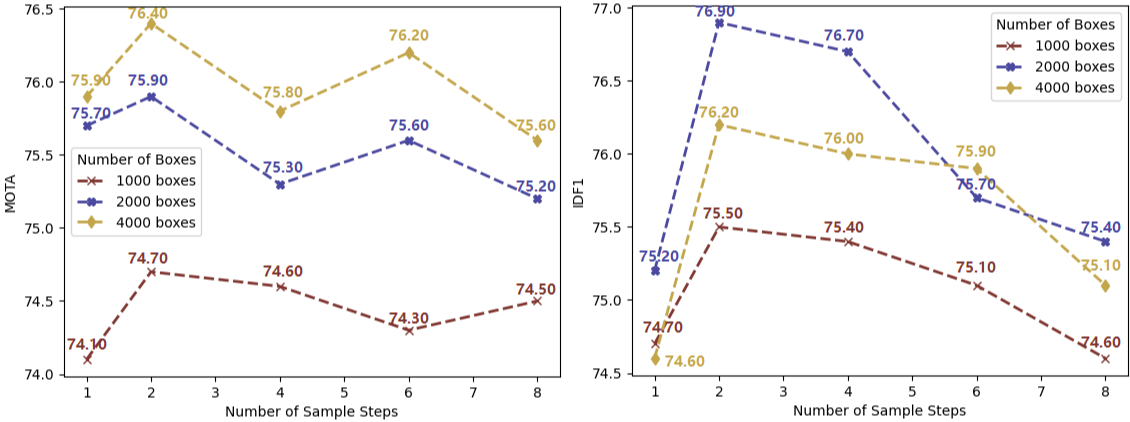}
		\caption{The performance of ConsistencyTrack is valuated on the MOT17 val-half set with different numbers of proposal boxes and different numbers of sampling time steps.
}
		\label{dmy2}
\end{figure*}

\textbf{Dynamic boxes.} Once the model is trained, it can be utilized by varying the number of boxes and the sampling time steps during inference. Consequently, a single ConsistencyTrack can be deployed across multiple scenarios, achieving the desired speed-accuracy trade-off without the need for retraining the network. In Fig. \ref{dmy2}, we evaluate ConsistencyTrack with 1000, 2000, and 4000 proposal boxes by increasing $n_{ss}$ from 1 to 8. The results indicate that highest MOTA in ConsistencyTrack can be achieved by increasing the number of random boxes. Moreover, the highest MOTA and IDF1 is achieved when $n_{ss} = 2$, this aligns with the few steps mapping characteristic of Consistency Model. Note that when $n_{ss} > 2$, the MOTA metric remains in a state of oscillating fluctuations, but its peak is lower than the case of $n_{ss} = 2$.

\subsection{Simulation Analysis}\label{subsec44}

\begin{figure*}[!ht]
		\centering
		\includegraphics[width=1.0\textwidth]{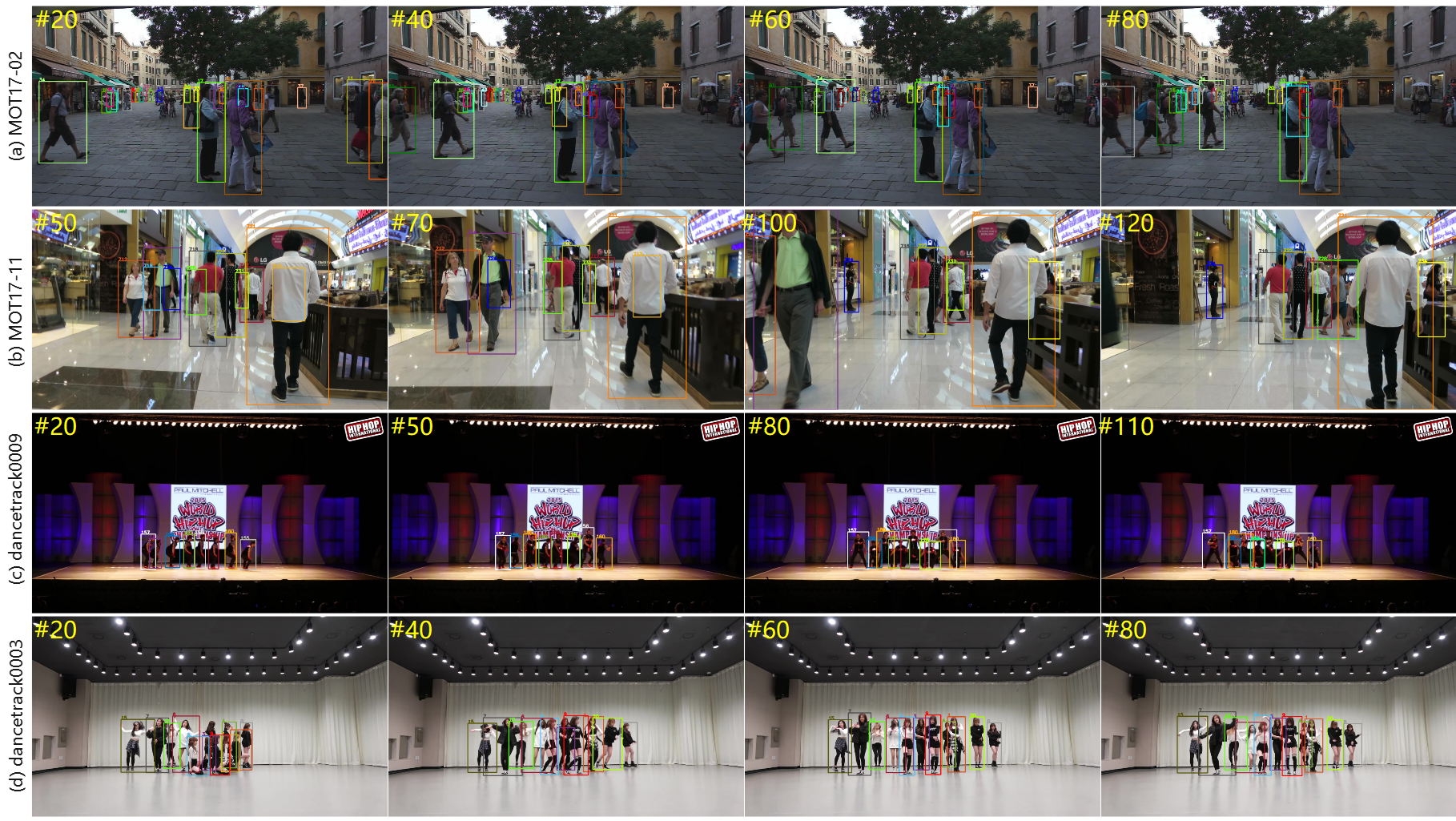}
		\caption{Screenshots of sampled tracking results on the proposed ConsistencyTrack on MOT17 and DanceTrack datasets. The frame numbers corresponding to the images are marked in the upper left corner.}
		\label{figtest}
\end{figure*}

The performance of ConsistencyTrack is evaluated against other tracking methods \cite{cai2022memot,zhou2020tracking,pang2020tubetk,cao2023retinamot,pang2021quasi} on the MOT17 and DanceTrack datasets. Tracking and matching results of ConsistencyTrack on the MOT17 and DanceTrack datasets are shown in Fig. \ref{figtest}. This subsection provides an analysis of the simulation results.

\textbf{MOT17 dataset.} The test performances on the MOT17 dataset are shown in Table \ref{mot17datasets}. The proposed ConsistencyTrack outperforms in several key metrics, achieving the best scores in MOTA (69.9\%), IDF1 (65.7\%), HOTA (54.4\%), and DetA (58.2\%). Compared to UTM, the proposed ConsistencyTrack improves MOTA and IDF1 by 6.4\% and 0.6\%, respectively. Additionally, ConsistencyTrack outperforms PCL in several metrics, such as IDF1, with an improvement of 4.5\%, demonstrating better tracking consistency and accuracy. ConsistencyTrack also excels in Mostly Tracked targets (MT) and Minimal Localization error (ML), with scores of 907 and 428, respectively, indicating strong capabilities in target location and its error control. All in all, compared to other methods, the proposed method shows significant advantages across multiple metrics, demonstrating that ConsistencyTrack offers superior overall accuracy and reliability.

\begin{table}[htbp]
  \centering
  \caption{Performance comparison on MOT17 test dataset on several metrics}
  \label{mot17datasets}
  \begin{adjustbox}{width=\textwidth}
  \begin{threeparttable}
  \begin{tabular}{l|ccccccccccc}
    \toprule
    \multirow{2}{*}{Methods} & \multicolumn{11}{c}{MOT17} \\
    \cmidrule(r){2-12}
    & MOTA$\uparrow$ & IDF1$\uparrow$ & HOTA$\uparrow$ & MT$\uparrow$ &   ML$\downarrow$ &  FP$\downarrow$   & FN$\downarrow$ &  AssA$\uparrow$ & DetA$\uparrow$ & IDs$\downarrow$ & Frag$\downarrow$ \\
    \midrule
    Tracktor++v2\cite{cai2022memot} & 56.3 & 55.1 & / & 498 & 831 & \textbf{8866} & 235449 & / & / & 1987 & /  \\
    TubeTK*\cite{pang2020tubetk} & 63.0 & 58.6 & 48.0 & 735 & \underline{468} & 27060 & 177483 & 45.1 & 51.4 & 4137 & 5727  \\
    CTTrack17\cite{zhou2020tracking} & \underline{67.8} & 64.7 & 52.2 & 816 & 579 & 18498 & \underline{160332} & 51.0 & \underline{53.8} & 3039 & 6102  \\
    CJTracker\cite{cao2023retinamot} & 58.7 & 58.2 & 48.4 & 621 & 909 & 32448 & 197790 & 48.0 & 49.1 & 2877 & 5031 \\
     TrajE\cite{girbau2021multiple} & 67.4 & 61.2 & 49.7 & 820 & 587 & 18652 & 161347 & 46.6 & 53.5 & 4019 &  6613 \\
     Sp\_Con \cite{wang2022split}  & 61.5 & 63.3 & 50.5 & 622 & 754 & 14056 & 200655 & \underline{52.0} & 49.2 & 2478 & 5079 \\
     PCL\cite{lu2024self} & 58.8  & 61.2 & 49.0 & 612 & 837 & \underline{12072} & 218912 & 51.1 & 47.2 & \textbf{1219} & \textbf{2197} \\
     UTM\cite{you2023utm} & 63.5  & \underline{65.1} & \underline{52.5} & \underline{881} & 635 & 33683 & 170352 & \textbf{53.2} & 52.2 & \underline{1686} & \underline{2562}   \\
    ConsistencyTrack & \textbf{69.9} & \textbf{65.7} & \textbf{54.4} & \textbf{907} & \textbf{428} & 24186 & \textbf{142145} & 51.2 & \textbf{58.2} & 3774 & 5854  \\
    \bottomrule
  \end{tabular}
  \begin{tablenotes}
    \item[1] Results of MOTA/IDF1/HOTA/AssA/DetA are percentage data (\%).
    \item[2] Bold font indicates the best performance while underlined font indicates the second best.
  \end{tablenotes}
  \end{threeparttable}
  \end{adjustbox}
\end{table}

\textbf{DanceTrack dataset.} In Table \ref{dancetrackdatasets}, we compared the ConsistencyTrack method with other traditional MOT methods on the DanceTrack validation set. Overall, our method demonstrated a balanced performance across various metrics and achieved a significant lead in the MOTA metric, reaching 88.1\%, far surpassing the metrics of the method in second place. Due to the inherent unfairness in comparing One-Stage strategy object tracking algorithms with those of other strategies, we chose to compare our algorithm with CenterTrack \cite{zhou2020tracking} and FairMoT \cite{zhang2021fairmot} on DanceTrack test set. The results are shown in Table \ref{dancetrackdatasets2}. The experimental results indicate that, except for the DetA metric, which is slightly lower than CenterTrack, all other metrics are higher than both CenterTrack and FairMoT.

\begin{table}[htbp]
  \centering
  \caption{Performance comparison on DanceTrack val set}
  \label{dancetrackdatasets}
  \begin{adjustbox}{scale=0.9}
  \begin{threeparttable}
  \begin{tabular}{l|ccccc}
    \toprule
    \multirow{2}{*}{Methods} & \multicolumn{5}{c}{DanceTrack} \\
    \cmidrule(r){2-6}
    & HOTA$\uparrow$ & DetA$\uparrow$ &AssA$\uparrow$ & MOTA$\uparrow$ & IDF1$\uparrow$ \\
    \midrule
    IoU\cite{cai2022memot} & 44.7 & \textbf{79.6} & 25.3 & \underline{87.3} & 36.8 \\
    DeepSORT\cite{wojke2017simple} & \textbf{45.8} & 70.9 & \textbf{29.7} & 87.1 & \textbf{46.8}  \\
    MOTDT\cite{chen2018real} & 39.2 & 68.8 & 22.5 & 84.3 & 39.6\\
    ConsistencyTrack & \underline{45.5} & \underline{77.7} & \underline{26.9} & \textbf{88.1} & \underline{43.4}  \\
    \bottomrule
  \end{tabular}
  \begin{tablenotes}
    \footnotesize
    \item[1] Results are all percentage data (\%).
    \item[2] Bold font indicates the best performance while underlined font indicates the second best.
  \end{tablenotes}
  \end{threeparttable}
  \end{adjustbox}
\end{table}

\begin{table}[htbp]
  \centering
  \caption{Performance comparison on DanceTrack test set}
  \label{dancetrackdatasets2}
  \begin{adjustbox}{scale=0.9}
  \begin{threeparttable}
  \begin{tabular}{l|ccccc}
    \toprule
    Methods & \multicolumn{5}{c}{DanceTrack} \\
    \cmidrule(r){1-6}
    One-Stage& HOTA$\uparrow$ & DetA$\uparrow$ &AssA$\uparrow$ & MOTA$\uparrow$ & IDF1$\uparrow$ \\
    \midrule
    CenterTrack\cite{zhou2020tracking} & 41.8 & \textbf{78.1} & 22.6 & 86.8 & 35.7 \\
    FairMOT\cite{zhang2021fairmot} & 39.7 & 66.7 & 23.8 & 82.2 & 40.8  \\
    ConsistencyTrack & \textbf{42.3} & 76.4 & \textbf{25.4} & \textbf{87.8} & \textbf{41.2}  \\
    \bottomrule
  \end{tabular}
  \begin{tablenotes}
    \footnotesize
    \item[1] Results are all percentage data (\%).
    \item[2] Bold font indicates the best performance.
  \end{tablenotes}
  \end{threeparttable}
  \end{adjustbox}
\end{table}

\subsection{Ablation studies}\label{subsec45}

Comprehensive ablation studies were conducted to elucidate the characteristics of the proposed ConsistencyTrack on the MOT17 val-half set. These simulations utilized YOLOX as the primary backbone, with no further modifications or enhancements specified.

\begin{table}[htbp]
  \centering
  \caption{Performance comparison with varied thresholds of Box-renewal and repeat times}
  \label{repeat_threshold}
  \begin{threeparttable}
    \begin{tabular}{@{}c@{\hspace{1cm}}c@{}}
      \begin{minipage}[t]{0.35\textwidth}
        \centering
        \resizebox{\textwidth}{!}{
        \begin{tabular}{cccc}
          \toprule
          $n_{rp}$ & MOTA$\uparrow$ & IDF1$\uparrow$ & IDP$\uparrow$ \\
          \cmidrule(r){1-4}
          6 & \underline{75.5} & \textbf{76.6} & \textbf{83.7} \\
          8 & \textbf{75.8} & 76.2 & \underline{82.9} \\
          10 & 75.2 & \underline{76.4} & 82.8 \\
          \bottomrule
        \end{tabular}}
      \end{minipage} &
      \begin{minipage}[t]{0.35\textwidth}
        \centering
        \resizebox{\textwidth}{!}{
        \begin{tabular}{cccc}
          \toprule
          $B_{th}$ & MOTA$\uparrow$ & IDF1$\uparrow$ & IDP$\uparrow$ \\
          \cmidrule(r){1-4}
          0.5 & \textbf{75.8} & 75.3 & 82.0 \\
          0.6 & \textbf{75.8} & \textbf{76.2} & \underline{82.9} \\
          0.7 & 75.4 & \underline{76.1} & \textbf{83.1} \\
          \bottomrule
        \end{tabular}}
      \end{minipage}
    \end{tabular}
    \begin{tablenotes}
      \footnotesize
      \item[1] Results are all percentage data (\%).
      \item[2] Bold font indicates the best performance while underlined font indicates the second best.
      \item[3] Left table represents the proportion of prior information results and right table represents Box-renewal results.
    \end{tablenotes}
  \end{threeparttable}
\end{table}
\textbf{Proportion of prior information.} In contrast to object detection, which operates without prior knowledge of object positions, MOT benefits from prior frame information. By incorporating this prior knowledge, we can adjust the proportion of such information in the construction of \(N_{\text{test}}\) proposal boxes by repeating the previous frame's boxes. In the experiments, we tested the impact of $n_{rp} \in \{1,2,\cdots,10\}$  on metrics such as MOTA, IDF1, and IDP, with the specific results for $n_{rp} = 6/8/10$ presented on the left of Table~\ref{repeat_threshold}. The results indicate that setting the Repeat parameter to 8, which involves repeating the prior boxes eight times from the \( (k-1) \)-th frame, yields the best performance.

\textbf{Box-renewal threshold.} The right column of Table~\ref{repeat_threshold} describes the performances with varied threshold $B_{th}$ on the metrics of MOTA, IDF1, and IDP. The case $B_{th}=0$ signifies that no threshold is applied. Analysis of the MOT17 validation set indicates that a threshold of 0.6 obtains a slightly better performance, compared to other thresholds.

\textbf{Accuracy vs. speed.} In Table~\ref{fps_table}, the inference speeds of ConsistencyTrack and DiffusionTrack are compared on the MOT17 val-half set. Operational efficiency was measured using a single NVIDIA RTX 3090 GPU with a batch size of one. The evaluation of DiffusionTrack is operated with varied sampling time steps ($n_{ss} = 2/4/6$) and a dynamic box count ($n_{p} = 2000$), and FPS is recorded. ConsistencyTrack was tested with steps of ($n_{ss}=1/2/4/6$) and a dynamic box count ($n_{p} = 2000$). Experimental results indicate that ConsistencyTrack not only achieves a sharp increase in FPS compared to DiffusionTrack under the same $n_{ss}$, but also maintains stable FPS as $n_{ss}$ increases incrementally, unlike DiffusionTrack. This demonstrates ConsistencyTrack's ability to significantly enhance inference speed.

\begin{table}[htbp]
  \centering
  \caption{Comparison of operational efficiency (FPS) between ConsistencyTrack and DiffusionTrack}
  \label{fps_table}
  \resizebox{0.5\textwidth}{!}{
  \begin{threeparttable}
    \begin{tabular}{@{}c|cc@{}}
      \toprule
      \multicolumn{1}{c|}{$n_{ss}$} & \multicolumn{2}{c}{FPS$\uparrow$} \\
      \cmidrule(lr){1-1} \cmidrule(lr){2-3}
       & DiffusionTrack & ConsistencyTrack \\
      \midrule
      1 & / & \textbf{10.53} \\
      2 & 2.50 & 10.51 \\
      4 & 1.25 & 10.39 \\
      6 & 0.84 & 10.27 \\
      \bottomrule
    \end{tabular}
    \begin{tablenotes}
      \item[1] Bold font indicates the best performance.
    \end{tablenotes}
  \end{threeparttable}
  }
\end{table}

\begin{table}[htbp]
  \centering
  \caption{Performance comparison of stretching methods on advanced metrics}
  \label{metrics_table}
  \resizebox{0.6\textwidth}{!}{
  \begin{threeparttable}
    \begin{tabular}{c|cccc}
      \toprule
      Stretching Method $f(x)$ & MOTA$\uparrow$ & IDF1$\uparrow$ & IDP$\uparrow$ \\
      \midrule
      $\frac{x - \mu}{\sigma}$ & 75.6 & 74.9 & 81.7 \\
      $e^x$ & 75.6 & \underline{75.7} & \underline{82.7} \\
      $\sqrt{x}$ & \underline{75.7} & 75.5 & 82.4 \\
      $\tanh(x)$ & \underline{75.7} & 75.4 & 82.2 \\
      $\log(x)$ & \textbf{75.8} & \textbf{76.2} & \textbf{82.9} \\
      \bottomrule
    \end{tabular}
  \begin{tablenotes}
    \footnotesize
    \item[1] Results are all percentage data (\%).
    \item[2] Bold font indicates the best performance while underlined font indicates the second best.
  \end{tablenotes}
    \end{threeparttable}
    }
\end{table}

\textbf{Stretch association function.}
In order to classify detection objects into high and low confidence levels, it is necessary to stretch data that is confined to a small range into a larger range. In Table \ref{metrics_table}, we investigated the impact of different stretching methods on the tracking performance. Specifically, we compared the effects of normalization stretching, exponential function stretching, square root function stretching, and hyperbolic tangent function stretching, and contrasted them with our logarithmic stretching method with a base of 1.01. All experiments were conducted under identical conditions, using $n_{p} = 2000$, $n_{ss} = 6$, $n_{rp} = 8$, and $B_{th} = 0.6$. The results indicate that, compared to the other methods, our approach demonstrates a distinct advantage in terms of MOTA, IDF1, IDP metrics.

\begin{table}[htbp]
  \centering
  \caption{Performance comparison between ConsistencyTrack and DiffusionTrack on basic metrics}
  \label{basic_metrics_table}
  \resizebox{0.8\textwidth}{!}{
  \begin{threeparttable}
    \begin{tabular}{cccccccc}
      \toprule
      Method & MOTA$\uparrow$ & IDF1$\uparrow$ & IDP$\uparrow$ & MT$\uparrow$ & ML$\downarrow$ & FN$\downarrow$ & IDs$\downarrow$\\
      \midrule
      DiffusionTrack & 74.4 & 74.5 & 82.7 & 46.6 & 10.6 & 21.3 & 433\\
      ConsistencyTrack & \textbf{75.7} & \textbf{76.5} & \textbf{83.3} & \textbf{52.8} & \textbf{18.6} & \textbf{19.4} & \textbf{298}\\
      \bottomrule
    \end{tabular}
  \begin{tablenotes}
    \footnotesize
    \item[1] Results of MOTA/IDF1/IDP/MT/ML/FN are percentage data (\%).
    \item[2] Bold font indicates the best performance.
  \end{tablenotes}
    \end{threeparttable}
    }
\end{table}

\begin{figure*}[ht]
		\centering
		\includegraphics[width=1.0\textwidth]{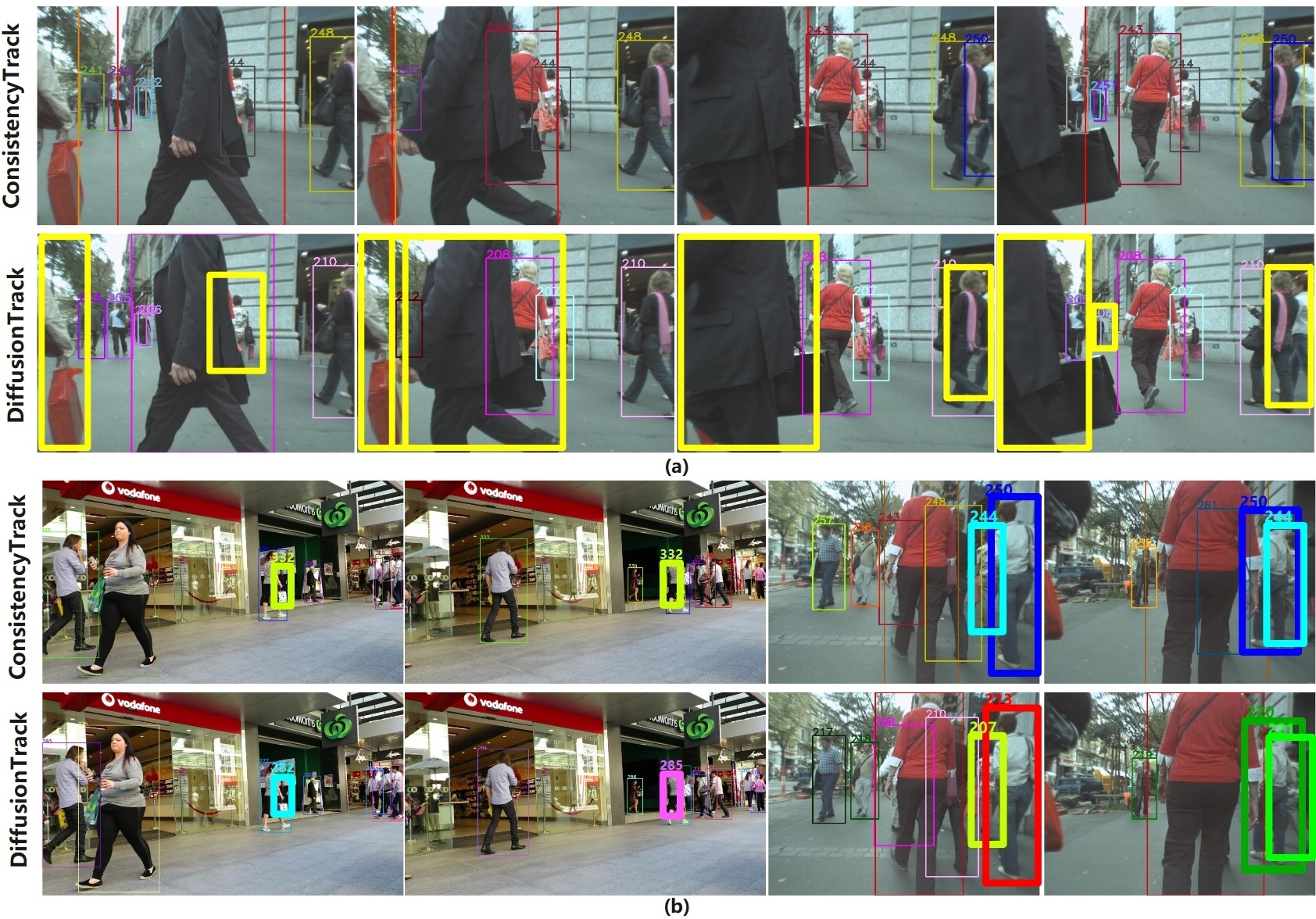}
		\caption{Performance comparison between ConsistencyTrack and DiffusionTrack on MOT17 val-half set.
  Fig. \ref{com2}(a) illustrates the robust performance of ConsistencyTrack when handling occlusions. The thicker yellow boxes highlight the areas where cases of incorrect detection occur. Fig. \ref{com2}(b) shows the cases that the exceptional ability of ConsistencyTrack in addressing the ID-switch problem after resolving occlusions. Bold boxes of the same color indicate the same ID.}
		\label{com2}
\end{figure*}

\textbf{New tracking and matching strategy.}
We validated the effectiveness of the tracking and matching strategy designed in this work, compared to DiffusionTrack. We set unified parameters, such as $n_p = 2000$, $n_{ss} = 6$, and $n_{rp} = 8$. The experiments showed improvements of 1.3\% in the MOTA, 2\% in IDF1, and 0.6\% in IDP. The visualization results are presented in Fig. \ref{com2}, which intuitively demonstrates the superior performance of the proposed ConsistencyTrack.

\begin{figure*}[!ht]
	\centering
	\includegraphics[width=1.0\textwidth]{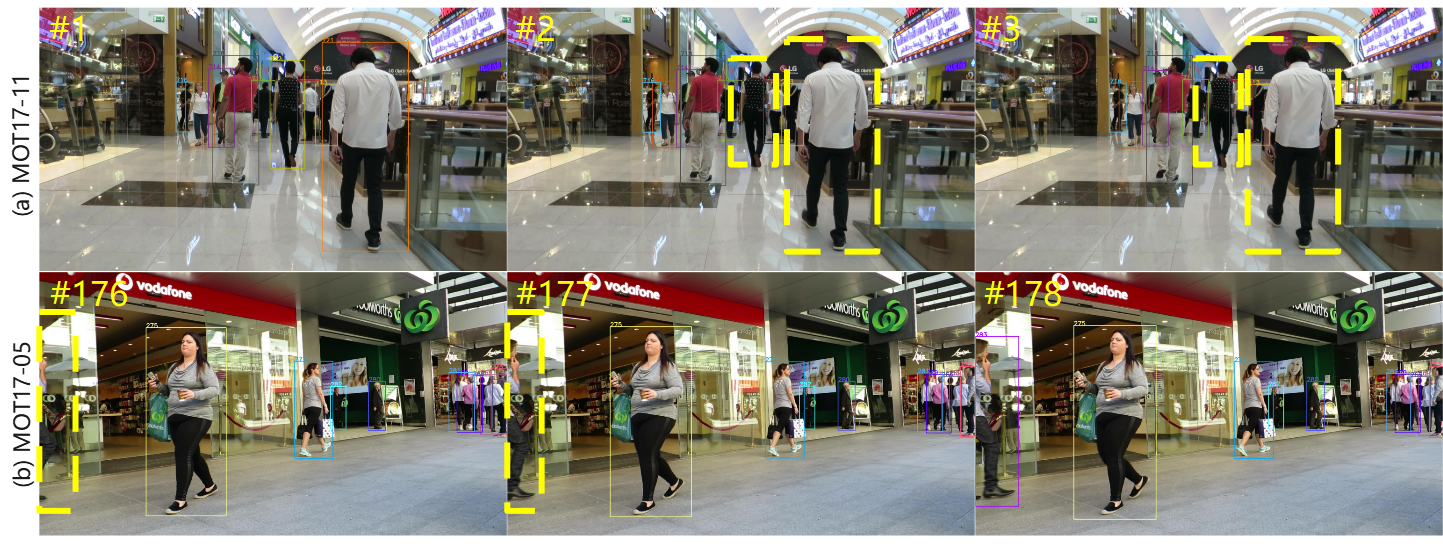}
	\caption{Tracking failures of the proposed ConsistencyTrack in MOT task. Few targets marked in yellow dashed bounding boxes are failed to be tracked continuously, with the situations of failed associations or missed detections with  partial occlusions.
}
	\label{an2}
\end{figure*}

In this section, the outstanding performance and prominent features of ConsistencyTrack are demonstrated on the MOT17 and DanceTrack datasets. Notably, this model significantly surpasses DiffusionTrack in terms of execution efficiency and maintains stability as the sampling timesteps increase, marking one of its most critical innovations. However, due to the model's reliance on very few denoising steps, a decline in accuracy is inevitable. This is primarily manifested in the frequent loss of tracked targets and delayed identification of new targets, with
typical failed cases presented in the Fig. \ref{an2}. These deficiencies still need to be enhanced with more perfect theoretical support
or the trade-off between tracking effect and efficiency.

\begin{figure*}[!ht]
		\centering
		\includegraphics[width=1.0\textwidth]{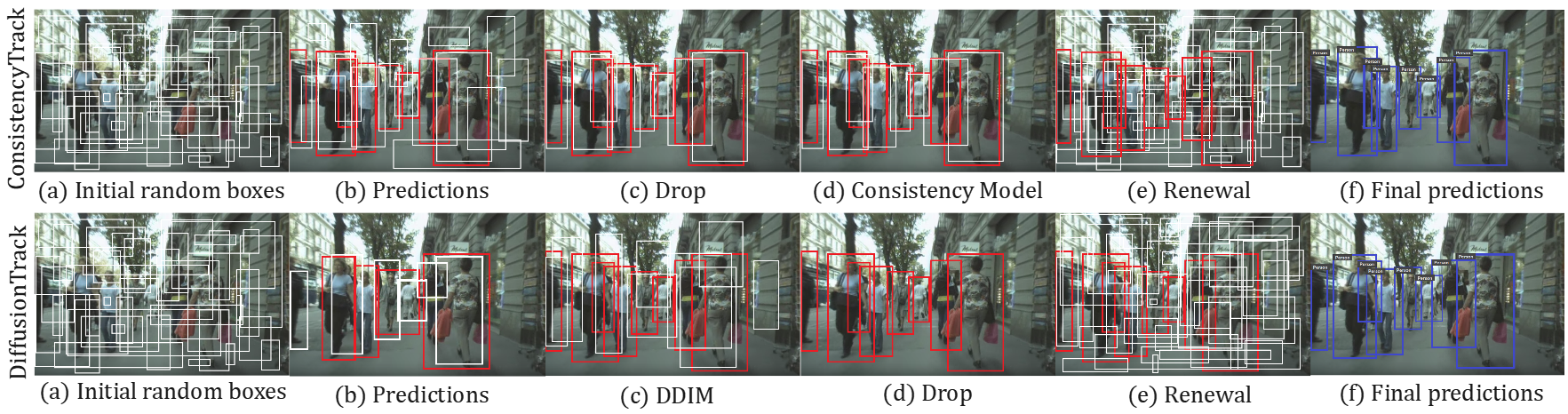}
		\caption{The comparison of the visual reasoning process with one typical sampling step between ConsistencyTrack and DiffusionTrack. The initial noised boxes or verified boxes with low confidence are marked in white, while the boxes with high confidence are marked in red and the final predictions are marked in blue.}
		\label{com1}
\end{figure*}

\section{Conclusions}\label{sec5}
In this work, we have introduced the generative principles of Consistency Model into an end-to-end MOT approach, implementing a JDT paradigm. Our noise-to-tracking pipeline possesses several attractive features, such as self-consistency and single-step denoising. The model's structure and unique self-consistency enable to achieve faster inference results with the same parameter settings. Extensive experiments demonstrate that ConsistencyTrack achieves excellent performance compared to previous methods. This work provides a novel insight into MOT from the perspective of Consistency Model and open up a new avenue in the field of MOT.

It is noteworthy that due to ConsistencyTrack's adoption of a single-step denoising method, its excessive noise addition and reduction amplitude has compromised its accuracy in MOT tasks. Future research will focus on enhancing the detection and tracking precision of ConsistencyTrack and exploring the way to integrate the core principles of Consistency Model into other advanced tracking models.




\clearpage
\bibliography{SCGTracker_original}

\end{document}